%% file: sm.tex
\documentclass[a4paper,fleqn]{cas-dc}

\usepackage[numbers]{natbib}

\begin{document}
\let\WriteBookmarks\relax
\def\floatpagepagefraction{1}
\def\textpagefraction{.001}
\shorttitle{Link Analysis meets Ontologies}
\shortauthors{S.M., M.B., N.L., B.\v{S}.}

\title[mode = title]{Link Analysis meets Ontologies: Are Embeddings the Answer?}

\author[1]{Sebastian Me\v{z}nar}[role=Researcher]

\credit{credit}

\author[1]{Matej Bevec}[role=Researcher]

\author[1,2,3]{Nada Lavra\v{c}}[%
   role=Researcher]
   
\author[1,2]{Bla\v{z} \v{S}krlj}[role=Researcher]

\address[1]{Jo\v{z}ef Stefan Institute, Jamova 39, Ljubljana, Slovenia}
\address[2]{Jo\v{z}ef Stefan International Postgraduate School, Jamova 39, Ljubljana, Slovenia}
\address[3]{University of Nova Gorica, Glavni trg 8, Vipava, Slovenia}

\credit{credit 2}

\begin{abstract}
The increasing amounts of semantic resources offer valuable storage of human knowledge; however, the probability of wrong entries increases with the increased size. The development of approaches that identify potentially spurious parts of a given knowledge base is thus becoming an increasingly important area of interest. In this work, we present a systematic evaluation of whether structure-only link analysis methods can already offer a scalable means to detecting possible anomalies, as well as potentially interesting novel relation candidates. Evaluating thirteen methods on eight different semantic resources, including Gene Ontology, Food Ontology, Marine Ontology and similar, we demonstrated that structure-only link analysis could offer scalable anomaly detection for a subset of the data sets. Further, we demonstrated that by considering symbolic node embedding, explanations of the predictions (links) could be obtained, making this branch of methods potentially more valuable than the black-box only ones. To our knowledge, this is currently one of the most extensive systematic studies of the applicability of different types of link analysis methods across semantic resources from different domains.
\end{abstract}


\maketitle

\section{Introduction}
\label{sec:introduction}

Researchers and companies often encounter tasks that, for their solution, need domain knowledge. This knowledge often contains complex relations between entities and human-defined terms that can be modelled using ontologies~\cite{brank2005survey, roche2003ontology}. Ontologies can range from small ones that describe people, their activities, and relations to other people (FOAF ontology~\cite{graves2007foaf}) up to large ones such as the Gene ontology that describes e.g., protein functions and cellular processes~\cite{Ashburner2000GO}. 

Ontologies have long been used to represent and reason over domain knowledge but have recently shown further potential in conjunction with machine learning methods. They have been used for relation prediction tasks~\cite{zhang2016ppi, chen2020candidategenes}, much like graphs, or to expand features with background knowledge in other machine learning tasks. Commonly applied methods range from more traditional semantic similarity approaches~\cite{pesquita2009semanticsim, zhang2016ppi} to recently successful entity embedding algorithms, whether it be graph-based~\cite{Grover2016node2vec, Burges2013TransE, lv2018transC, chen2018on2vec}, syntactic~\cite{smaili2018onto2vec, smaili2018opa2vec}, or hybrid~\cite{chen2020owl2vec}.
Alternatively, ontologies have been utilized to constrain the output of machine learning and optimization models to conform to certain rules~\cite{silla2011hiercls}. Machine learning has also been used to help experts with ontology construction tasks. \textit{Ontology matching} refers to finding a semantic mapping between two inter operable ontologies~\cite{doan2004ontomatch}, while \textit{ontology completion} refers to finding missing, non-present links that are not logically deducible from the existing ontology but are plausible~\cite{chen2020owl2vec, li2019ontocomp}. 

While small ontologies can easily be annotated due to a limited number of possible relations, even highly skilled domain experts can make mistakes when considering larger ontologies, by missing some links or adding non-existent ones. These mistakes can have an impact on our understanding of the domain and can produce models and solutions that do not perform well. In our work, we aim to adopt machine learning methods that perform well on the link prediction task to find the missing and the redundant links in ontologies solely based on their structure. We do this by representing ontologies as graphs, and applying link prediction methods. The contributions of this work can be summarized as follows:
\begin{itemize}
    \item A proposed methodology for finding missing and redundant links in real-life ontologies.
    \item Demonstrated utility of the considered link analysis methods on multiple ontologies with different properties.
    \item Simple to use software for prediction of missing and redundant links and evaluation of the proposed methodology on a given ontology.\footnote{Found at \url{https://github.com/smeznar/link-analysis-meets-ontologies}}
    \item Proposed temporal approach for evaluating the quality of the methods, investigating how the predicted links are accounted for in the future versions of the same ontology.
    \item An investigation of how link predictions can be \emph{explained} via symbolic node embedding.
\end{itemize}

\section{Related work}
In this section, we first introduce ontologies and their use in machine learning, followed by the relevant work from the field of anomaly detection, machine learning on graphs and link prediction.

\subsection{Ontologies}
Ontologies refer to machine-readable representations of knowledge in a given application domain, usually defined in a declarative knowledge modelling language, such as OWL (Web Ontology Language)~\cite{w3c2012owl} that is based on description logic (DL). Ontologies operate with individuals, classes (sets of individuals), and properties (relations between individuals), for which they define semantics through a set of logical statements -- axioms.

These statements fall into two categories. The terminological box (also called the T-box, vocabulary, or schema) contains statements defining classes, their characteristics and hierarchy. In contrast, the assertional box (A-box) consists of assertions about individuals (concrete facts), which use the vocabulary of the T-box. Given a complete ontology, reasoners can infer additional implicit facts from the explicitly defined set based on rules defined in the T-box. For example, the A-box fact \textit{"Mary is a mother"} implies \textit{"Mary is a parent"} since the T-box defines that \textit{"mother is a subclass of parent"}.

Some of the T-box statements in OWL are class subsumption axioms (\textit{"mother SubclassOf: woman"}, meaning that every mother is a woman), property restrictions (\textit{"parent EquivalentTo: HasChild Some person"}, meaning that everyone who has at least one person as a child is a parent) and set operators (\textit{"parent EquivalentTo: mother Or father"}, meaning that a parent is either a mother or a father).  A-box statements include membership axioms (\textit{"Mary Types: mother"} meaning Mary is a mother) and property assertions (\textit{John Facts: HasWife Mary}, meaning that Mary is John's wife). These examples are taken from W3C's OWL primer~\cite{w3c2012owl} and are presented in the Manchester syntax, a notation that will also be used in further examples.

Ontologies are used in various fields and vary significantly in their purpose, content, and implementation. However, they correspond to one of the two archetypes outlined below.

\begin{itemize}
\item 
The first refers to small, ungrounded ontologies that lack an A-box and serve as a semantic schema of high-level terms (classes) in a particular domain. They are often used in the scope of the semantic web and are intended to be referenced by various web sources and thereby populated with "external" facts. Examples include FOAF (12 classes) and Marine ontology (104 classes). In the former, the classes represent agents, documents, and other entities on the web, while the latter describes high-level terms relating to marine biology (see Section~\ref{sec:datasets}).

\item Conversely, there are larger, more grounded ontologies attempting to comprehensively capture knowledge in a domain as a complex hierarchy with many concrete facts. An illustrative example is the Gene ontology~\cite{Ashburner2000GO} with (at the time of writing this paper) 62{,}201 classes representing gene products and their functions.
\end{itemize}

Notably, such ontologies, or at least their graph representations, could be considered knowledge graphs by some definitions that define the latter as a schema (T-box) accompanied by a large number of (A-box) facts~\cite{paulheim2016kg, bonatti2018kg}. However, knowledge graphs do not have a well-established definition yet, with other work giving alternate proposals, such as a collection of facts without the schema~\cite{kejriwal2020kg2}, or a system encompassing an ontology and a reasoning engine~\cite{ehrlinger2016kg3}.

In practice, different grounded ontologies take different approaches to capturing A-box (ground level) facts in OWL. Some, such as HeLiS~\cite{dragoni2018helis}, use individuals and OWL's object property assertion axioms to define that two "ground level entities" are related by a property. Others, such as the Gene ontology and Food ontology~\cite{Dooley2018FoodOn}, do not to use individuals at all, representing even very "individual-like" entities as classes and thus blurring the line between T-box and A-box or between an ontology and a knowledge graph (KG). Instead, such ontologies define the equivalent of a property assertion with a construct like the following (from FoodOn):

\begin{quote}
\textit{pecan pie SubclassOf: HasIngredient Some sugar}
\end{quote}

Directly, this expresses that \textit{"pecan pie is a subclass of the class of all dishes, which have some sort of a sugar as an ingredient"}, but is intended to be read as \textit{"pecan pie has sugar as an ingredient"}.

\subsection{Mining ontologies with network-based approaches}

As described, ontologies have recently shown potential when used with machine learning methods, such as to provide additional background information or to constrain the learning process, but they can also be seen as standalone resources, much like knowledge graphs.
In our work, we attempt to systematically approach mining ontologies with graph-based methods to identify anomalies and potentially novel relations.

The related work comprises of several papers that present ontology-specific embedding algorithms.
Onto2Vec~\cite{smaili2018onto2vec} and OPA2Vec~\cite{smaili2018opa2vec} constructs sentences from OWL axioms and trains a language model, On2Vec~\cite{chen2018on2vec} \linebreak is based on translational graph embeddings and \linebreak OWL2Vec*~\cite{chen2020owl2vec} combines the language model approach with random walks on the ontology graph. These methods are typically evaluated against knowledge graph embeddings on a limited number of large ontologies in a relation prediction task.

Other examples are some approaches to semantic data mining~\cite{Kralj2019NetSDM} and domain-specific applications, most often in the biomedical domain, where ontologies are mined for tasks such as protein-protein interaction (PPI) prediction and gene function prediction, gene-disease prediction. Here several ontologies are usually combined into a single data set. As for methods, semantic similarity approaches~\cite{zhang2016ppi, zhao2018genefunc}, often heavily tailored to the task, are the traditional choice, but many recent works adopt ontology-specific embeddings~\cite{nunes2021genedisease, chen2020candidategenes, althubaiti2019cancergenes}.

None of the above could be considered a systematic study. Perhaps the closest work to ours and a valuable resource about the topic is a recent survey on the state of machine learning with ontologies~\cite{kulmanov2020ontosurvey} that covers both traditional semantic similarity methods and recent embedding-based methods. It looks at simple graph embeddings, knowledge graph embeddings and ontology-specific embeddings, categorizing them into graph-based, syntactic and semantic approaches. The study also includes an experimental evaluation of a subset of these methods in a protein-protein interaction task. However, it only considers two subsets of GO as data sets, since its focus is on theoretic categorization of the field and on the biomedical domain in particular.

Comparatively, some key characteristics of our study are:
\begin{itemize}
    \item We evaluate a large number of graph-based methods and compare simple graph embeddings to KG-specific embeddings. However, we limit outselves to structure-only methods due to the nature of the ontology-to-graph conversion.
    \item We test our methodology on substantially different ontologies both in terms of the domain they cover and their size, ranging from small schemas to large knowledge bases of ground level facts.
    \item We propose a temporal approach for evaluating the quality of the methods.
\end{itemize}

A more extensive overview of related work is summarized in Table \ref{tab:related-work}.

 \begin{table*}[pt!]
    \centering
    \let\oldstrech\arraystretch
      \renewcommand{\arraystretch}{1.7}
    \caption{Overview of some of the related approaches, their aim and a short description. The horizontal lines separate different types of research articles.}
	 \resizebox{.95\textwidth}{!}{
	\input{table1}
	}
	\label{tab:related-work}
	\let\arraystretch\oldstrech
\end{table*}

\subsection{Graph-based machine learning}
Graph-based machine learning has seen a rise in popularity in recent years due to its potential to work with complex data structures such as relational databases and structures commonly found in biology and chemistry~\cite{costa2011analyzing}. This branch of machine learning mainly focuses on node and graph classification~\cite{Bhagat2011nodeclass}, node clustering, and link prediction tasks~\cite{lu2011link}.

Machine learning tasks on graphs are usually solved in one of three ways. Traditionally, tasks on graphs were solved using label propagation~\cite{zhu2002learning}, PageRank~\cite{ilprints422}, and proximity-based measures such as Adamic/Adar~\cite{Adamic2003adamic}, Jaccard coefficient~\cite{Salton1983introduction}, and preferential attachment. Another group of approaches embed graphs into tabular data which is used together with traditional machine learning methods such as logistic regression to generate predictions. These approaches include well-established and tested methods such as \linebreak node2vec~\cite{Grover2016node2vec} and Deepwalk~\cite{Perozzi2014deepwalk}, as well as some new ones such as SNoRe~\cite{Meznar2020snore}. Recently, with new research in deep learning approaches, neural network models such as graph convolutional networks (GCN)~\cite{kipf2017semi}, and graph attention networks (GAT)~\cite{velickovic2018graph} have emerged as end-to-end learners.

\subsection{Anomaly detection}
Anomaly detection studies patterns in data and searches for instances that do not conform to them. These instances usually represent anomalies, cases that need further examination to stop malicious attempts or give additional information about the observed environment. Anomaly detection can be found in many different domains such as intrusion detection~\cite{Denning1987Intrusion}, fraud detection~\cite{Fawcett99activitymonitoring}, medical anomaly detection~\cite{Horn2002Effect}, and image protection~\cite{Pokrajac2007LocalOutliers}.

Detection methods usually depend on the domain of focus and its data. The concept of an anomaly is not well defined, so what might be considered an anomaly in the medical field might not be considered an anomaly in fraud detection due to different nature of the data. Broadly, anomalies can be split into point anomalies, contextual anomalies, and collective anomalies. Point anomalies are those where each instance can be considered an anomaly regarding other data. Contextual anomalies are the ones considered anomalous in some context but might not be otherwise. An example of a contextual anomaly is the high temperature during winter. Lastly, collective anomalies are a collection of instances that are anomalous in regards to other data, i.e. one second of readings during a electrocardiogram scan~\cite{Chandola2009Anomaly}.

Anomalies can be found using different approaches. One approach is to use classifiers that output whether an instance is anomalous or not. The selection of the classifier does not matter much, but neural networks, Bayesian networks, and support vector machines have been mostly used in recent years. Other approaches include statistical and information theoretic-based detection, nearest-neighbour based detection, where we assume anomalies lie far from their neighbours, clustering-based detection, where we assume anomalies lie outside clusters, while non-anomalous data lie inside. The last approach is spectral-based detection and embeds instances into latent space where anomalies differ significantly from non-anomalous data.

The proposed approach is closely related to contextual anomaly detection and link prediction tasks. The distribution is, in our case, the link structure of the ontology; an anomaly is a connection that exists (or is missing) but should not, based on the structure of the ontology. The contextual part of the approach comes from the fact that some connections might exist in one context (one domain) but not in a different one. Most algorithms we use first create embeddings of a network's nodes, subsequently used in classification. These algorithms closely resemble those that are used in spectral-based anomaly detection. This methodology is explained in more detail in Section~\ref{sec:methodology-anomaly}.

\subsection{Link prediction}
Link prediction is one of the most widely addressed tasks concerning network-based data. Predicting whether there exists an edge between two nodes without any additional information is almost impossible, but with some additional information about the network, the nodes, and with some assumptions, various approaches predict such existence well~\cite{liben2007link,lu2011link}. 

The most common assumption used in link prediction is that two nodes are connected if they are similar. This similarity might be due to them sharing similar node features or having common neighbours (or, more commonly, neighbourhoods). When applied to real-life networks, this assumption is very reasonable since, for example, two individuals who went to the same school, lived in the same city, and had similar hobbies are more likely to know each other and thus be friends on social media networks. Another assumption that is commonly used is that nodes are likely to be connected to nodes with a high number of neighbours. This also mirrors real-life networks as, for example, a paper with many citations is more visible and thus more likely to be cited by a new paper than a paper only cited once.

Link prediction is traditionally solved using proximity-based methods that model the networks using the assumptions mentioned above. These methods most commonly predict the links based on the first and second neighbours of their nodes, e.g., the number of common neighbours. These include the Jaccard index~\cite{Salton1983introduction}, Adamic/Adar index~\cite{Adamic2003adamic}, and others.

With the rise of graph-based machine learning, new methods were developed that usually perform very well on all networks, even the ones where the discussed (structure) assumptions do not hold. These methods mostly embed nodes into a dense low dimensional, or sparse representation that maintains the network's structure. The most common such embedding method is node2vec~\cite{Grover2016node2vec}, which uses random walks as sentences for training the skip-gram model~\cite{Mikolov2013word2vec}. Another approach, SNoRe~\cite{Meznar2020snore}, creates a sparse embedding where a node is represented as a vector of similarities between the hashed neighbourhood of the node and the neighbourhoods of nodes selected as features.

Similar to embedding methods, graph neural networks have recently been used for different machine learning tasks on real-life networks These approaches jointly exploit the adjacency matrix of a network alongside node features. The most popular graph neural network approaches include graph convolutional networks (GCN)~\cite{kipf2017semi}, graph attention networks (GAT)~\cite{velickovic2018graph} and graph isomorphism networks (GIN)~\cite{xu2018powerful}.

For \emph{knowledge graphs}, i.e., graphs, where nodes and edges usually contain some additional information (mostly types), specialized approaches can be used. One such approach is metapath2vec~\cite{dong2017metapath2vec} that works similarly to node2vec, but the sampled random walks have a predetermined structure (meta paths). Other approaches such as TransE~\cite{Burges2013TransE} and RotatE~\cite{sun2018rotate} embed nodes and relations in such a way that a combination of their embeddings creates a vector that has a norm close to zero if the triplet (subject, predicate, object) is inside the graph and close to one if it is not.

\section{Methodology}

In this research, we aim to represent ontologies as graphs and employ link prediction methods to find missing and redundant connections. The proposed methodology consists of the following steps: ontology to graph transformation, link prediction, and finding of missing and redundant connections described below.

\subsection{Ontology to graph transformation}\label{sec:ontology-transformation}

As mentioned in Section~\ref{sec:introduction}, machine learning has approached the use of ontologies in various ways, such as treating them as documents~\cite{smaili2018opa2vec, smaili2018onto2vec}. One common approach is to represent ontologies as graphs where nodes represent classes or individuals, and links encode semantic relationships defined by the ontology. This addresses some limitations of existing non-graph methods~\cite{chen2020owl2vec, kulmanov2020ontosurvey}, but more importantly, it enables the use of many powerful graph-based machine learning methods that are being developed for other problem domains and are rapidly evolving.

Since an ontology can be understood as a set of logical expressions and is usually modelled as such, there exist multiple possible conversions of a given ontology into a graph. Certain expressions, like property assertions between individuals (facts) and class taxonomy (\textit{"subclass of"} relations) directly map to nodes and edges. Others, like property restrictions, domain-range axioms, and set operators, do not have an obvious representation. Given our aim to learn about ontologies using graph-based methods, the conversion needs to be such that semantics expressed with OWL axioms are sufficiently reflected in the resulting graph's topology.

A number of different approaches to converting OWL ontologies to graphs have been developed for various tasks. However, there is not yet an established standard or agreement on what the most appropriate representation is.
W3C provides a specification for translating the OWL syntax directly to a heterogeneous graph represented by RDF triples~\cite{w3c2012owl2rdf}. Conversely, most other methods attempt to approximate semantic relations between entities by introducing new links that are somehow justified by the ontology's axioms. This branch of methods includes Onto2Graph~\cite{rodriguez2018onto2graph}, OBO Graphs~\cite{smith2005relations} and so-called projection rules~\cite{soylu2017projection, chen2020owl2vec}.

We consider two of these approaches: the OWL to RDF-triplet mapping following the W3C specification and conversion by projection rules. We choose the former because it is a W3C standard and the latter because it has been used in a similar machine learning context. However, we note that these conversion algorithms could be substituted with others without significant changes to the rest of our methodology.

\begin{itemize}
\item
The first approach \cite{w3c2012owl2rdf} exactly reproduces the OWL file in a graph (RDF triple) form, including all axioms without any reductions. Simple relations, such as  (\textit{"pecan pie SubclassOf: food product"}), are transformed directly. Complex expressions, like property restrictions, are translated by introducing so-called \emph{blank nodes}. For example, the previously mentioned FoodOn construct \textit{"pecan pie SubclassOf: HasIngredient some sugar"} is transformed into 4 triples: \textlangle{}\textit{pecan pie, SubclassOf, x}\textrangle{}, \textlangle{}\textit{x, SomeValuesFrom, sugar}\textrangle{}, \textlangle{}\textit{x, type, Restriction}\textrangle{}, \textlangle{}\textit{x, OnProperty, HasIngredient}\textrangle{}. 
In theory, this means all the expressiveness is kept, and a reasoner can traverse such a graph, inferring new facts. However, in a machine learning setting, where such rigour is not necessary, all the intermediate syntactic nodes and edges might only act as noise when learning associations among the entities of interest.
\item
The second strategy is based on so-called projection rules proposed in~\cite{soylu2017projection} and has been used for machine learning with ontologies~\cite{chen2020owl2vec}, producing good results. Here, not all exact logical relationships are kept. Simple relations like class subsumption and property assertions between individuals are again transformed directly. However, complex logical expressions, like property restrictions, are approximated with simple triples. The same example \textit{"pecan pie SubclassOf: HasIngredient some sugar"} becomes the intuitive triple \textlangle{}\textit{pecan pie, HasIngredient, sugar}\textrangle{}, meaning \textit{pecan pie} and \textit{sugar} are directly connected by an edge, without any intermediate nodes.
The result is a graph that presumably at least approximately captures all relationships but doesn't contain noisy syntactic structures.
\end{itemize}

Since our work focuses on methods that only consider graph structure, data properties (like \textit{"Mary has age 35"}) and other lexical information like labels and descriptions of entities (annotation properties) are discarded in the conversion.

As described, both conversion methods produce a directed heterogeneous multigraph, i.e. a set of triples (edges) of the form $\langle{}s, p, o\rangle{} \in T$, where $s$ and $o$ are nodes (classes, individuals, or blank nodes) and $p$ is a label, representing the relation between them.
This is the input to our chosen knowledge graph embedding methods that operate on heterogeneous graphs. For other methods, we further convert this graph into an undirected simple graph $G(N,E)$ so that $\{o,s\} \in E \Leftrightarrow \exists p: \langle{}o,p,s\rangle{} \in T$, meaning two nodes are at most connected by a single undirected anonymous edge.

Previous work has taken different approaches, like including all available lexical information along with graph topology and training language models to produce embeddings~\cite{chen2020owl2vec}.

In our experiments, we found conversion by projection rules to outperform the OWL to RDF mapping across the board. We, therefore, choose to only present results using graphs obtained with conversion by projection rules (see Section \ref{sec:results-link}).

\subsection{Link prediction benchmark}\label{sec:methodology-link}
Link prediction and anomaly detection tasks are closely related since the anomaly detection model must be able to reconstruct the graph to be able to predict if an edge (connection) between two nodes is an anomaly or not. Because of this, it is crucial to determine how well our model works on the link prediction task. High accuracy on the link prediction task means that the model will be able to reconstruct the graph well and thus accurately predict which edges (connections) are missing or redundant. 
In our work, we use the following methodology to test how well our methods perform on the link prediction task. First, we transform an ontology into a graph as described in Section~\ref{sec:ontology-transformation}.

After this, we create positive (existent) and negative (non-existent) examples. We shuffle and split them into five folds. We then use the edges from four out of five folds to create the adjacency matrix used in training.

\begin{figure*}[b!]
  \centering
\includegraphics[width=0.8
\linewidth]{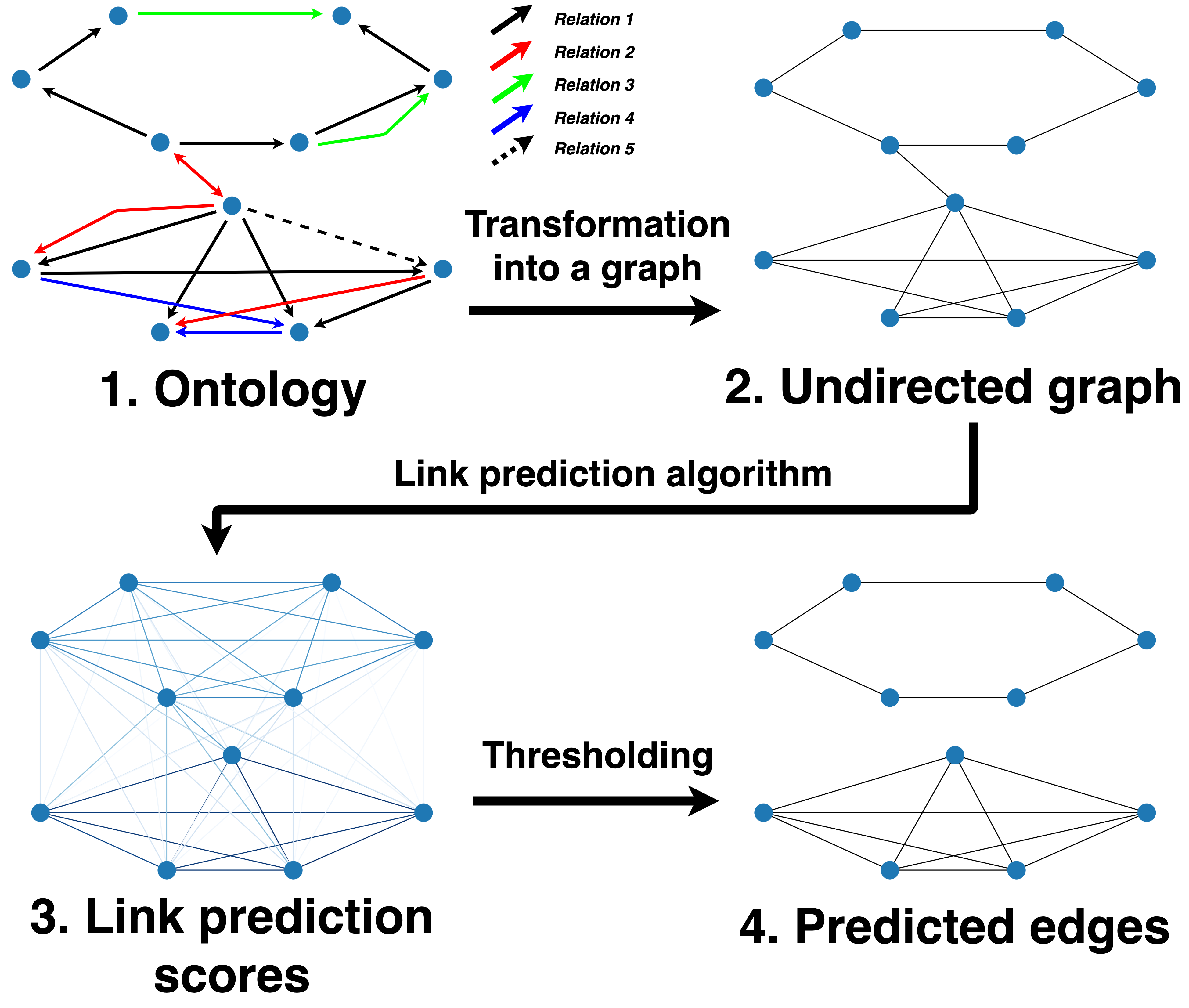}
  \caption{Overview of the link prediction methodology. We start with an ontology in stage 1 and transform it into an undirected graph to get to stage 2. Using link prediction algorithms, we get to stage 3, the link prediction scores. Lastly, we threshold these scores to get the predicted edges (stage 4).}
  \label{fig:overview}
\end{figure*}

For each fold, we then train the baseline models using the adjacency matrix generated from the other four folds and the corresponding positive and negative edges, if they are needed as input. We use these models to predict the existence of the positive and negative edges in this fold. We evaluate the performance using the ROC-AUC and average precision. An overview of the link prediction process can be seen in Figure~\ref{fig:overview}.

\subsection{Finding missing and redundant edges}\label{sec:methodology-anomaly}
Annotations of data are not perfect, and often an annotator might miss some relations in the ontology, or sufficient experiments might not have been conducted to determine if some relation exists. Because of this, methods for finding and recommending missing or redundant edges help improve the ontology. We create recommendations for such edges in the following way.

\begin{figure*}[b!]
  \centering
\includegraphics[width=0.8
\linewidth]{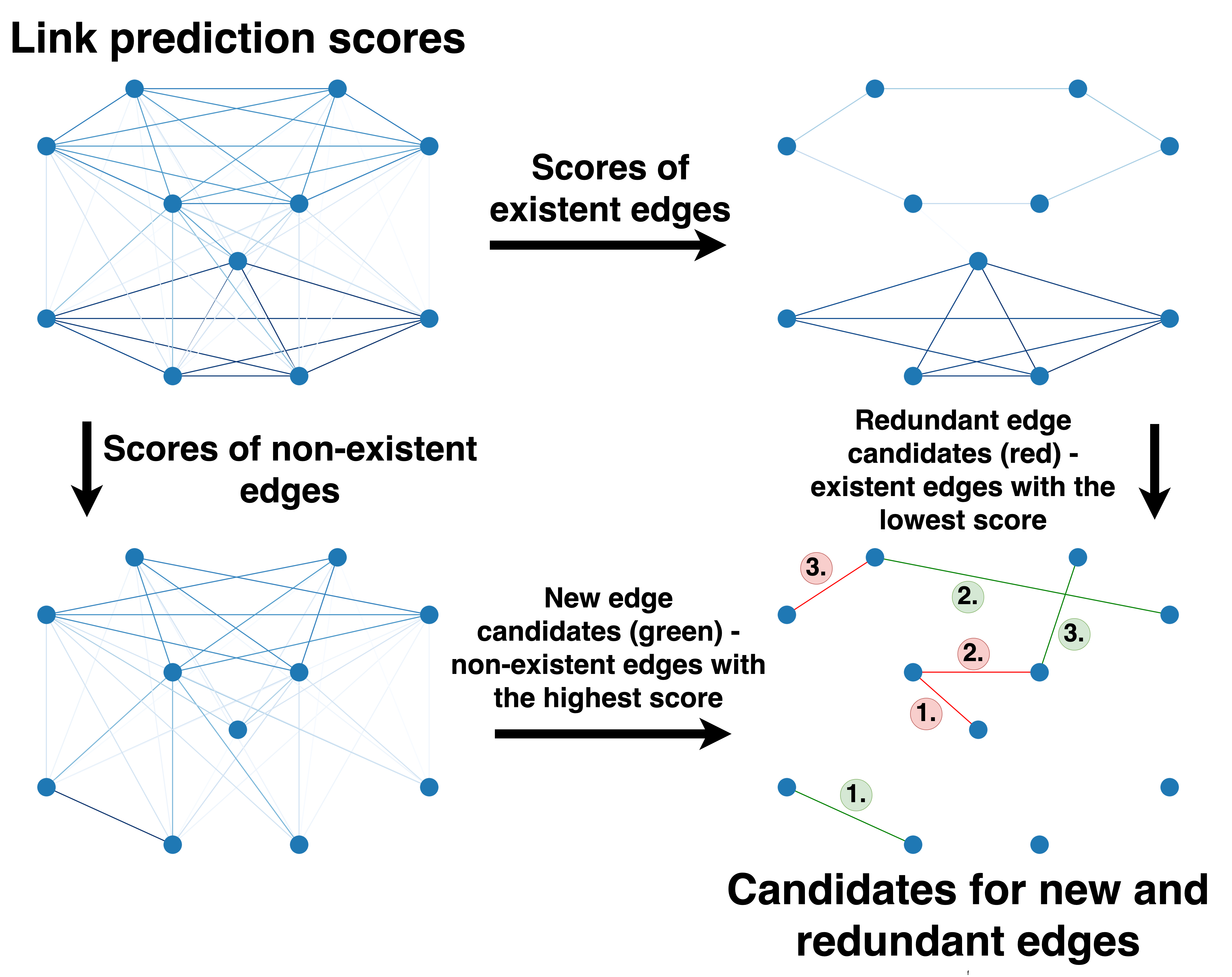}
  \caption{Overview of our methodology for finding missing and redundant edges. We start by splitting the link prediction scores into scores of the existent edges and those of non-existent edges. We take the non-existent edges with the highest score (candidates for missing edges) and existent edges with the lowest score (candidates for redundant edges) as our candidates.}
  \label{fig:overview-prediction}
\end{figure*}

We first transform the ontology as presented in Section~\ref{sec:ontology-transformation}. Then we embed the generated network into matrix $\boldsymbol{R}$ using SNoRe with the default parameters, i.e. $1024$ walks of maximum length $5$ for each node, inclusion threshold $0.005$, cosine similarity as the distance metric, and less than $|N|\cdot 256$ non-zero values. Using $\boldsymbol{R}$ we create the link prediction matrix $\boldsymbol{L} = \boldsymbol{R} \cdot \boldsymbol{R}^T$ that is used to find candidates for the missing and redundant connections. We split the link prediction matrix into two matrices, one that represents the score of existing edges and one that represents the score of non-existing edges. The matrix for the scores of existing edges can be obtained by using the adjacency matrix as the mask, while the matrix with scores for the non-existing edges can be obtained by subtracting the matrix with scores of existing edges from the link prediction matrix $\boldsymbol{L}$. Recommendations for non-existing edges are obtained by selecting the elements in the matrix with scores of the non-existing edges with the highest scores. Similarly, recommendations for existing edges are obtained by selecting elements with the lowest score in the matrix with scores of existing edges. An overview of this methodology is shown in Figure~\ref{fig:overview-prediction}. In the figure, the graph in the top left represents the $K_n$ graph weighted by the values in matrix $\boldsymbol{L}$. The edge scores are then split into those belonging to existent edges and non-existent edges. We chose the non-existent edges with the highest score (green) as candidates for missing edges and existent edges with the lowest score (red) as candidates for redundant edges.

Because the space complexity of the link prediction matrix is quadratic, such an approach might not be feasible for ontologies with many entities. One way to avoid this is to create predictions only for a small number of nodes $P\subset V$. This can be done by creating a prediction matrix $\boldsymbol{L}[P] = \boldsymbol{R}[P]\cdot \boldsymbol{R}^T$, where $\boldsymbol{A}[B]$ represents rows of nodes from $B$ in matrix $\boldsymbol{A}$. To obtain recommendations, we use the same technique as before, the only difference being that we use only the mask for the selected nodes. For the approaches that do not generate an embedding, this is done by only generating scores for the selected pairs of nodes (a subset).

\begin{figure*}[b!]
  \centering
\includegraphics[width=0.8
\linewidth]{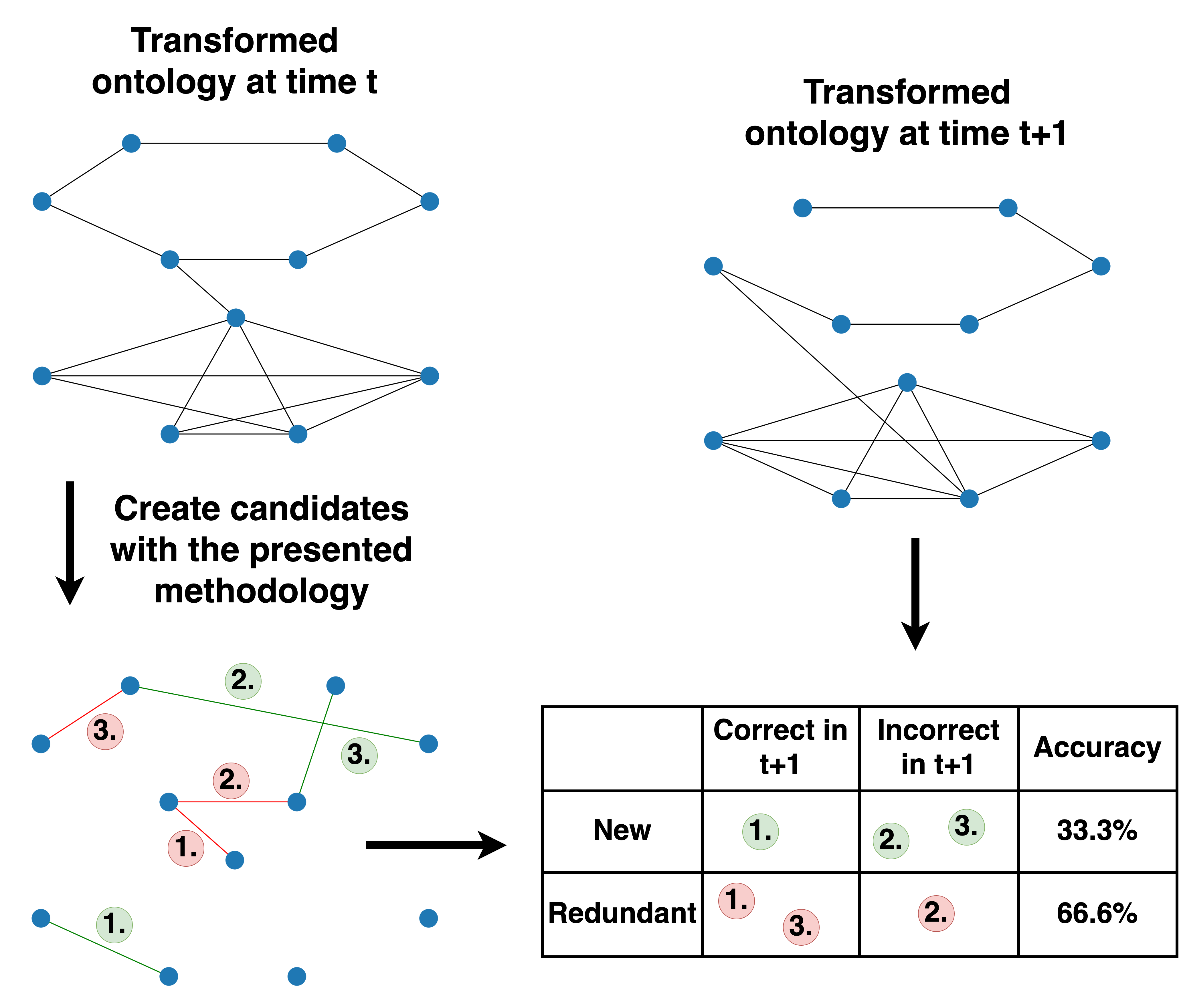}
  \caption{Overview of our temporal-based benchmark. We start with two transformed ontologies published at time $t$ and $t+1$. We create candidates for missing and redundant edges on the ontology published at time $t$ and test them on the ontology published at time $t+1$. The final score is obtained by dividing the number of correctly classified candidates by the total number of candidates.}
  \label{fig:overview-scoring}
\end{figure*}

We benchmark our approach for the missing and redundant edges using a temporal approach. We take different versions of one ontology, order them by their release date, and evaluate our methodology on them using the following approach shown in Figure~\ref{fig:overview-scoring}. We take two subsequent ontologies, one published at time $t$ and the other at time $t+1$, and transform them. Then we create candidates for missing and redundant edges on the ontology published at time $t$ and test whether these candidates occur in the ontology published at time $t+1$. If a candidate for the missing edge occurs in the ontology published at time $t+1$ we classify the candidate as correct, otherwise as incorrect. Similarly, we classify the candidate for the redundant edge as correct if the edge does occur in ontology published at time $t$ but not in the one published at time $t+1$. Otherwise, the candidate is classified as incorrect. We test different numbers of top $k$ candidates and get the score (accuracy) for two subsequent years as the number of correctly classified candidates is divided by the total number of candidates.

\subsection{Explaining recommendations}\label{sec:methodology-explanation}
In Section~\ref{sec:methodology-anomaly} we create recommendations for missing and redundant links using SNoRe. Besides good performance, SNoRe creates an embedding with symbolic features, which make the embedding and thus the recommendations easier to interpret. The following sections present our methodology to create both local interpretations (for a single recommendation) as well as global ones (which features contribute most to the connectivity of an ontology).

\subsubsection{Global interpretation}
Using global interpretations, we can further explore the domain and how much the selected (symbolic) features influence the existence of a link between two nodes. The importance of a global interpretation can be given with the toy example where nodes represent pages (people and interests) on a social network and links represent the connection between two pages (friendship or a follow). Let us say that the similarity to the neighbourhood of a node representing a famous person and a local high school are selected as the relevant features. Since two people are more likely to know each other if they went to the same high school, the global interpretation will tell us that the feature representing the similarity to the neighbourhood of the local high school node is more important when determining the friendship between two people.

We create a global interpretation using logistic regression in the following way. First we create the node embedding using SNoRe and select edges that will be used as positive and negative examples during the training (similarly as in the link prediction benchmark~\ref{sec:methodology-link}). After this we create training data by multiplying the embeddings of nodes incident to the selected edges element-wise. We use this data to train the logistic regression model. The importance of features can be estimated by the absolute value of its t-statistic~\cite{molnar2019}. The t-statistic is calculated as the weight of the feature divided by its standard error. For $j$-th feature with the weight $\beta_j$, the t-statistic is calculated using the formula 
\begin{equation*}
    \textsc{SE}(\beta_j)=\sqrt{(\boldsymbol{X}^T\boldsymbol{W}\boldsymbol{X})^{-1}_{jj}},
\end{equation*}
where $\boldsymbol{X}$ represents the matrix with training data used as input to the logistic regression, $\boldsymbol{W}$ is a diagonal matrix where 
\begin{equation*}
    \boldsymbol{W}_{jj} = \frac{e^{\sum_{i=1}^p \beta_i \boldsymbol{X}_{ji}}}{(1+e^{\sum_{i=1}^p \beta_i \boldsymbol{X}_{ji}})^2},
\end{equation*}
and $p$ the number of features in the embedding.

\subsubsection{Local interpretation}
Usually, we are more interested in the local interpretations of the recommendations. These give us an insight into what contributed to the prediction of an edge between two nodes. Let us show an example of such interpretation with the same toy scenario as for the global interpretation. In this scenario, one user might be recommended to another one due to the high value of the local high school feature, but another one due to the high value of the feature representing the similarity to the neighbourhood of the famous person node.

We create a local interpretation by multiplying the embeddings of two nodes incident to the selected edge element-wise. We can interpret the recommendation by sorting features based on their value and looking at which features contributed the most to the confidence score for that edge. Further, if we use a classifier such as logistic regression, features can be weighted using the parameters of that model.

\section{Experimental setting}
In this section, we first present the considered data sets (ontologies), followed by description of the baselines and the evaluation procedures.

\subsection{Data sets}\label{sec:datasets}
In our work, we tested ontologies of different sizes and with different properties. These ontologies are the following: 

\begin{description}
\item[Marine TLO]\cite{tzitzik2013marineTLO}
    A small top-level ontology of concepts related to biodiversity data in the marine domain. It is intended to help integrate new information about marine species (linked data) by providing a hierarchy of generic classes like \textit{legislative\ zone} or \textit{ecosystem}.
\item[Anatomy ontology (AEO)]\cite{Bard2012AEO}
    A high-level vocabulary of anatomical structures common across species. It aims to enable interoperability between different anatomy ontologies (such as EHDAA2) and describes anatomical entities such as \textit{artery}, \textit{bone} or \textit{mucous membrane}.
\item[SCTO]\cite{El-Sappagh2018SCTO}
    Captures upper-level terms from the Systematized Nomenclature of Medicine (SNOMED CT), a comprehensive medical terminology used to manage electronic health data, as an ontology. It is just a taxonomy (only \textit{subclass\ of} relations), with diverse classes such as \textit{symptom}, \textit{laboratory\ test} or \textit{anatomical\ structure}.
\item[Emotion ontology (MFOEM)]\cite{Hastings2011Emotions}
    It aims to describe affective phenomena (emotions and moods), their different building blocks, and their effects on human behaviour (expressions). Similar to the Anatomy ontology, this ontology includes more numerous and more specific terms than FOAF or SCTO, but is not as grounded as, for example, the Gene Ontology. It models classes such as \textit{anxiety}, \textit{negative\ valence} or \textit{blushing} and properties such as \textit{has\ occurrent\ part}.
\item[Human Developmental Anatomy v2 (EHDAA2)]\cite{Bard2012EHDAA2}
     An ontology that is primarily structured around the parts of organ systems and their development in the first 49 days (Carnegie stages (CS)1–20). It includes more than 2000 anatomical entities (AEs) and aims to include as much information about human developmental anatomy as is practical and as is available in the literature. 
\item[Food ontology (FOODON)]\cite{Dooley2018FoodOn}
    It aims to name all parts of animals, plants, and fungi that can bear a food role for humans, as well as derived food products and the processes used to make them. It is a large, fairly grounded ontology with upper-level entities like \textit{part\ of\ organism} and leaf classes as specific as \textit{Pinot noir wine} or \textit{chickpea}. Some properties include \textit{has\ ingredient}, \textit{derives\ from} and \textit{has\ quality}.
\item[LKN]\cite{Ramsak2018lkn} 
    A biological \emph{knowledge graph} constructed from multiple different sources of information, including temporal expression data, small RNA-based interactions and protein-protein interactions. This source was obtained in the process of semi-automatic curation.
\item[Gene ontology (GO)]\cite{Ashburner2000GO}
    A comprehensive source of information on cellular processes. It describes three types of entities - molecular functions, cellular components and biological processes - and their relations in a complex class hierarchy linked mostly by \textit{is\ a} (subsumption), \textit{part\ of} (meronymy) and \textit{regulates} properties. Among the used ontologies, GO is the largest and most grounded with entities (classes) ranging from \textit{molecular function} to, for example, \textit{DNA alpha-glucosyl transferase activity}.
\end{description}

Some basic statistics of these ontologies are shown in Tables~\ref{tab:ontology-statistics} and~\ref{tab:onto_stats}. We have four smaller ontologies, one medium sized, and three bigger ones. Of the three bigger ontologies, the Food ontology has a very tree-like structure, while LKN and gene ontologies are more connected. We also ran more extensive and automated experiments on different versions of the Gene ontologies between the years 2015 and 2021.

\begin{table}[]
    \centering
    \input{table2}
    \caption{Basic statistics of the tested ontologies' graph forms (by projection rules), where |N| denotes the number of nodes and |E| the number of edges.}
    \label{tab:ontology-statistics}
\end{table}

\begin{table*}[htb!]
    \centering
	\input{table3}
	\caption{Basic statistics of the tested OWL ontologies.
	Set axioms encompass class equivalence, disjointness, union and intersection.
	Imported ontologies are considered part of the importing ontology --- this applies to Emotion, Food and Gene ontology.
	}
	\label{tab:onto_stats}
\end{table*}

\subsection{Baselines}
We test our approach with the following baselines:

\begin{description}
    \item[Adamic/Adar]\cite{Adamic2003adamic} An edge between nodes $u$ and $v$ is scored with the formula $\sum_{x \in N(u) \cap N(v)} \frac{1}{\log |N(x)|}$ where $N(x)$ is the neighborhood of node $x$. These scores are normalized and thresholded to obtain link prediction. 
    \item[Jaccard coefficient]\cite{Salton1983introduction} An edge between nodes $u$ and $v$ is scored with the formula $\frac{|N(u) \cap N(v)|}{|N(u) \cup N(v)|}$, where $N(u)$ is the neighborhood of node $u$. These scores are normalized and thresholded to obtain link prediction.
    \item[Preferential attachment]\cite{Barabasi1999preferential} An edge between nodes $u$ and $v$ is scored with the formula $|N(u)|\cdot |N(v)|$, where $N(u)$ is the neighborhood of node $u$. These scores are normalized and thresholded to obtain link prediction.
    \item[GAE]\cite{kipf2016variational} Generates a node representation with a variational graph auto-encoder that uses latent variables to learn an interpretable model.
    \item[GAT]\cite{velickovic2018graph} Includes the attention mechanism that helps learn the importance of neighboring nodes. In our tests, we adapt the implementation from PyTorch Geometric~\cite{Fey2019ptg}.
    \item[GCN]\cite{kipf2017semi} A method that introduced convolution to graph neural networks and revolutionized the field. This approach aggregates feature information from the node’s neighborhood. In our tests, we adapt the implementation from PyTorch Geometric~\cite{Fey2019ptg}. 
    \item[GIN]\cite{xu2018powerful} Learns a representation that can provably achieve the maximum discriminative power. In our tests, we adapt the implementation from PyTorch Geometric~\cite{Fey2019ptg}.
    \item[SNoRe]\cite{Meznar2020snore} A node embedding algorithm that produces an interpretable embedding by calculating the similarity between vectors generated by hashing random walks. 
    \item[node2vec]\cite{Grover2016node2vec} A node embedding algorithm that learns a low dimensional representation of nodes that maximizes the likelihood of neighborhood preservation using random walks.
    \item[metapath2vec]\cite{dong2017metapath2vec} A node embedding algorithm that learns a low dimensional representation of nodes. The algorithm works similarly to node2vec, but samples random walks based on predetermined metapaths.
    \item[TransE]\cite{Burges2013TransE} creates a knowledge graph embedding in such way that the distance between the embedding of the second node and the embedding of the first node translated by the embedding of the relation is small.
    \item[RotatE]\cite{sun2018rotate} creates a knowledge graph embedding. This approach is similar to TransE, but instead of translating the embedding of the first node by the embedding of the relation, it rotates the embedding in complex vector space.
    \item[Spectral clustering]\cite{Ng2001onspectral} Generates a node embedding by using a non-linear dimensionality reduction method based on spectral decomposition of the graph Laplacian matrix. 
\end{description}

\subsection{Evaluation}
We evaluate the link prediction capabilities on transformed ontologies by using five-fold cross-validation. We create these folds as follows. We start with a directed (multi)graph with multiple edges between each pair of nodes. We transform this graph into a simple undirected graph and remove elements on the diagonal of the adjacency matrix (self-loops). Afterwards, we take the upper triangle of the adjacency matrix, put the elements into an array, and shuffle them to create positive examples. This is crucial for fair evaluation since each edge is chosen exactly once and thus contained inside exactly one fold. For negative examples, we randomly sample pairs of nodes, test if the edge between them exists, discard them in this case, and make sure they do not repeat. We use the same amount of positive and negative examples. We split positive and negative examples into five equally sized parts (last being a few examples shorter if the number of edges is not divisible by five).

We get the score of a baseline on the selected data set by taking the mean value of scores for each fold. A fold is scored by training the model with data from other folds and using either ROC-AUC or average precision to obtain prediction scores for edges in this fold.

To predict the edge score for the TransE and RotatE baseline, we also need to input the relation we want to predict. To bypass this, we generate predictions for each relation and output the most probable one. We do this because if there is no edge between two nodes, all predictions should have a low score, and otherwise, at least one should have a high score (the one we select).

We run our tests on a machine with 64GB of RAM and 12 threads. To make our experiments reproducible, we initialize the random number generators of the data splitting algorithm and all baselines with a predetermined seed. This way, data splits are the same for each baseline.  

\section{Results}\label{sec:results}
In this section we present the results of link prediction, results of temporal approach presented in section~\ref{sec:methodology-anomaly}, and show examples of explanations for the generated recommendations. 

\subsection{Link prediction results}\label{sec:results-link}
The results of link prediction using the ROC-AUC metric using the methodology described in Section~\ref{sec:methodology-link} are presented in Table~\ref{tab:roc}. From the results we can observe two aspects that generally hold for all baselines: the variance of results falls with the number of edges, and that baselines perform significantly worse on ontologies where the ratio between nodes and edges is close to one. We can also observe that on smaller ontologies embedding methods that do not rely on random walks such as spectral embedding, TransE, and RotatE work best, while on bigger ones SNoRe and TransE generally outperform the others. By grouping baselines of similar kinds together we see that proximity-based approaches usually give mediocre performance, graph neural networks work well on most data sets but usually fall just below the best performing approaches, node embedding algorithms based on random walks generally perform great on all data sets but are near the top on larger ones, approaches designed for knowledge graphs perform similarly to other embedding methods, and spectral embedding generally performs better on smaller ontologies, the exceptions being Marine and LKN. Overall, the best performing baselines are SNoRe and TransE.

\begin{table*}[htb!]
    \centering
    \resizebox{\textwidth}{!}{
	\input{table4}}
	\caption{Link prediction results based on the ROC-AUC metric (multiplied by 100 to improve readability).}
	\label{tab:roc}
\end{table*}

Similar results can be observed when the average precision metric is used. These results are presented in Table~\ref{tab:ap}. We can see that the baselines perform better on bigger graphs where the ratio between edges and nodes is not close to one. The biggest difference in performance using this metric can be observed on the GAE baseline that achieves the best score on the SCTO and EHDAA2 ontologies and is close the the best one on a few others. Another notable result is the performance of the baseline RotatE on Anatomy, when the baseline performs the worse with the ROC-AUC metric, but the best when average precision is used.

\begin{table*}[htb!]
    \centering
    \resizebox{\textwidth}{!}{
	\input{table5}}
	\caption{Link prediction results based on the average precision metric (multiplied by 100 to improve readability).}
	\label{tab:ap}
\end{table*}

\begin{figure*}[h!]
  \centering
  \includegraphics[width=\linewidth]{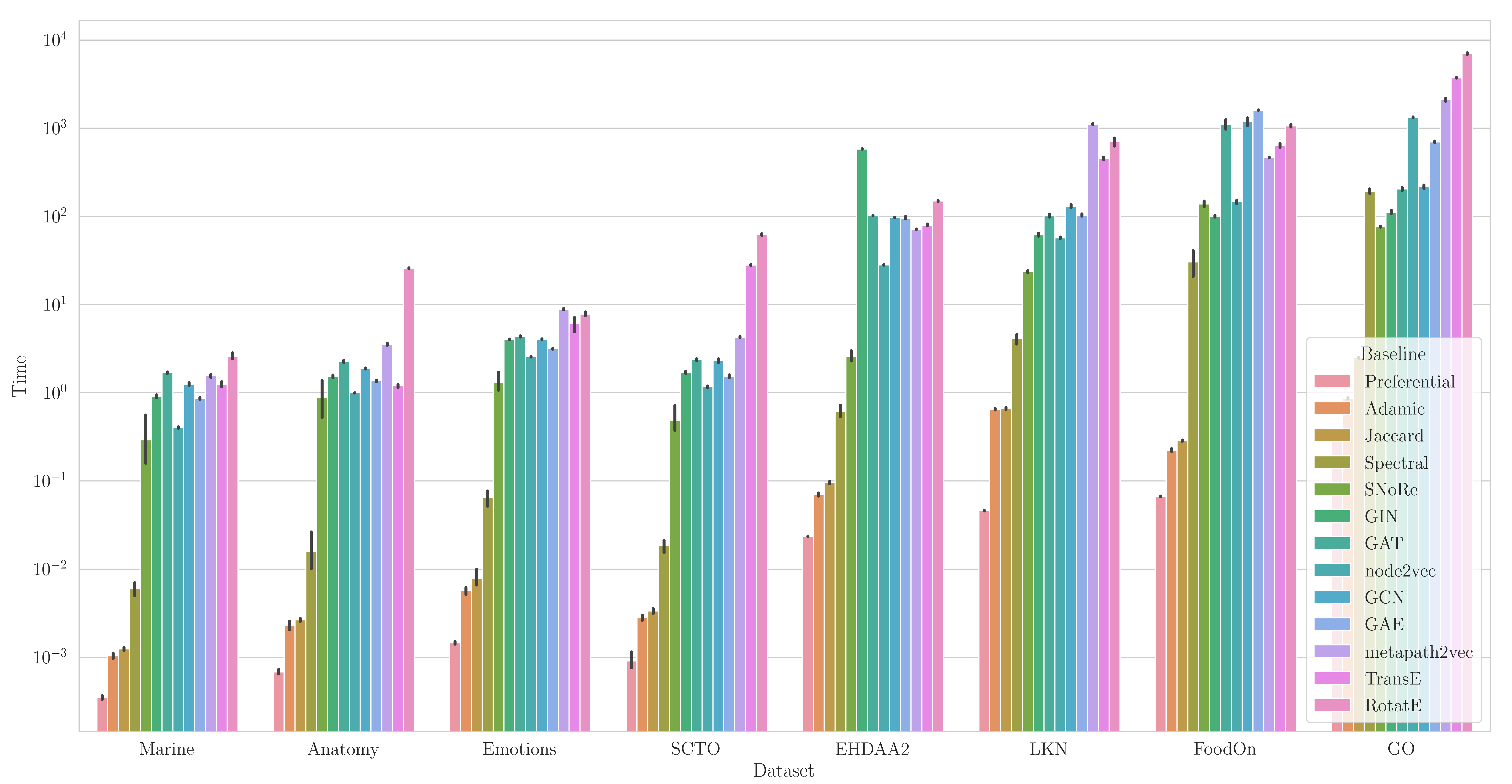}
  \caption{Time needed to train the baselines, shown on the logarithmic scale. Baselines are ordered by the total training time}
  \label{fig:time}
\end{figure*}

Lastly, Figure~\ref{fig:time} shows the average time needed for link prediction on a given ontology. We can see that the running time of smaller ontologies is less than a second for (almost) any given baseline. On the three bigger ontologies, proximity-based methods are the fastest even though they sometimes achieve great results, see for example the Preferential attachment on LKN. Overall, the slowest method is RotatE that needs almost twice as much time on the Gene ontology than TransE that is the second slowest. We can see that graph neural network methods perform similarly or a bit slower than SNoRe, while achieving lower results overall. 

These results show that link prediction performance and thus also the presented anomaly detection approach works best on ontologies that have enough nodes and more importantly edges.

\subsection{Detection of missing and redundant edges}\label{sec:results-candidates}
We tested our methodology for finding missing and redundant edges using the temporal approach described in Section~\ref{sec:methodology-anomaly} on the Gene ontology. We did this by selecting seven ontologies published between 2015 and 2021. The number of edges that were added and removed in each year is shown in Table~\ref{tab:gene-added-removed}. We can see that on average there are around 8400 edges added and 7500 edges removed each year.

We generated a different number of candidates (10, 100, 500) for both missing and redundant edges and tested how many of them were added to or removed from the ontology of the following year.

\begin{table*}[htb!]
    \label{tab:gene-added-removed}
    \centering
	\input{table6}
	\caption{Number of edges added to and removed from the previous year's Gene Ontology.}
\end{table*}

The accuracy of top-k missing candidates added in the following year is shown in Table~\ref{tab:gene-missing}. We can see that the candidates with a high score are more likely to be added in the following year. This is best shown on predictions generated on the ontology from $2019$, where $3$ of the top $10$ generated candidates were added, but then only $5$ more in the next $90$ and $18$ in the predictions in the range $100-500$.

\begin{table*}[htb!]
    \centering
	\input{table7}
	\caption{Accuracy of top k predicted missing edges at year $t$ that appear in the following year's ($t+1$) Gene Ontology.}
	\label{tab:gene-missing}
\end{table*}

The accuracy of top-k redundant candidates added in the following year is shown in Table~\ref{tab:gene-redundant}. These results are a bit different, mainly for $k=10$ where only one predicted edge gets removed in the following year. Results for $k=100$ and $k=500$ are very similar to those of the missing edges.

\begin{table*}[htb!]
    \centering
	\input{table8}
	\caption{Accuracy of top k predicted redundant edges at year $t$ that were removed in the following year's ($t+1$) Gene Ontology.}
	\label{tab:gene-redundant}
\end{table*}

\subsection{Interpretation examples}
To further make our predictions useful, we interpret them using the methodology presented in section~\ref{sec:methodology-explanation}.

\begin{figure*}[h!]
  \centering
  \includegraphics[width=\linewidth]{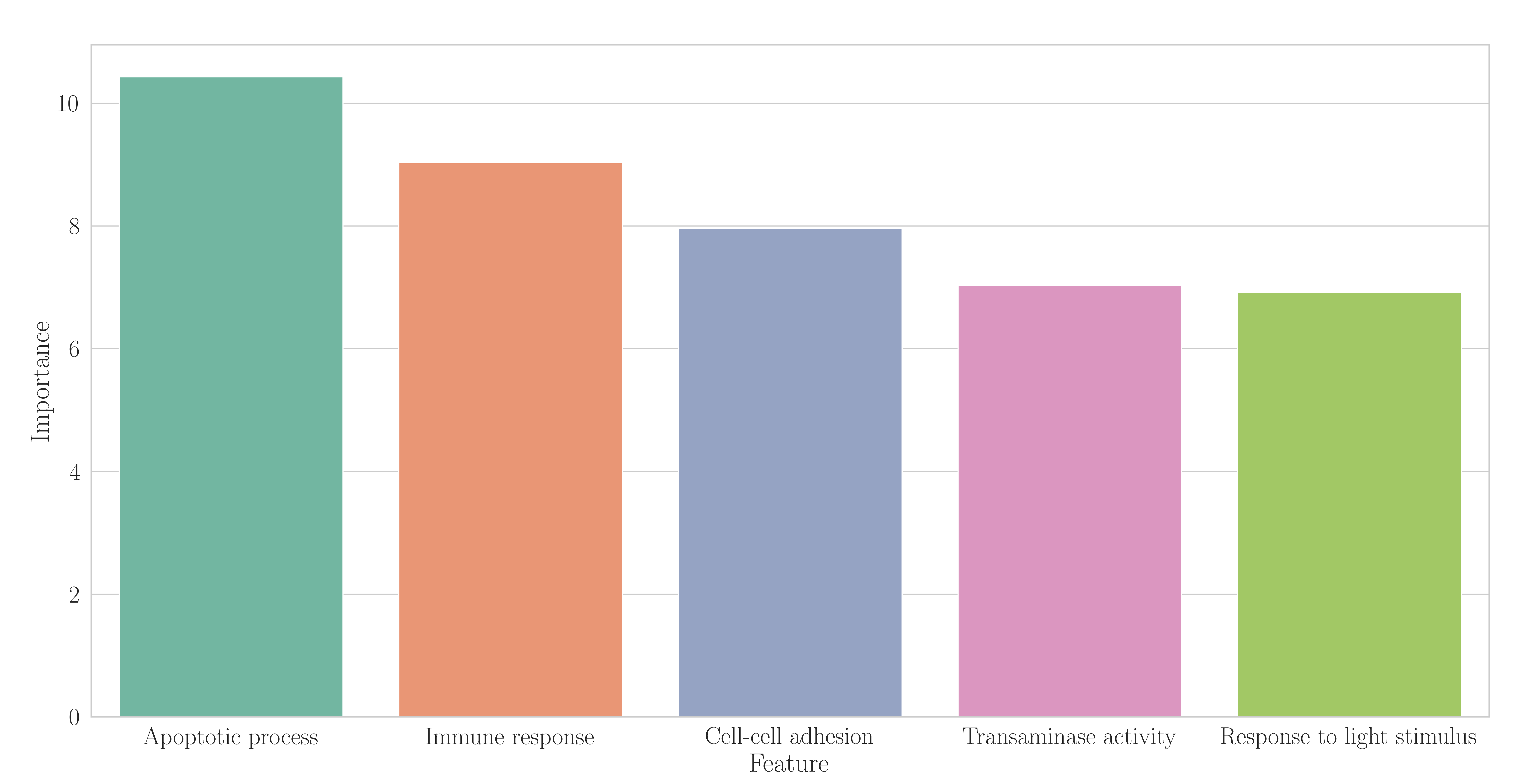}
  \caption{An example of a global explanation for the Gene ontology.}
  \label{fig:exp-global}
\end{figure*}

Figure~\ref{fig:exp-global} shows the global interpretation of recommendations for the 2019 gene ontology. In the figure, we show the ten features whose parameters have the highest values. We can see that the parameter of the term \emph{apoptotic process} has the highest value, meaning that when the neighborhood of two nodes is similar to the neighborhood of this node (high value in the embedding for this feature), it is more likely that there is an edge between them. Other features have lower parameter value but their value is still quite high.

\begin{figure*}[h!]
  \centering
  \includegraphics[width=\linewidth]{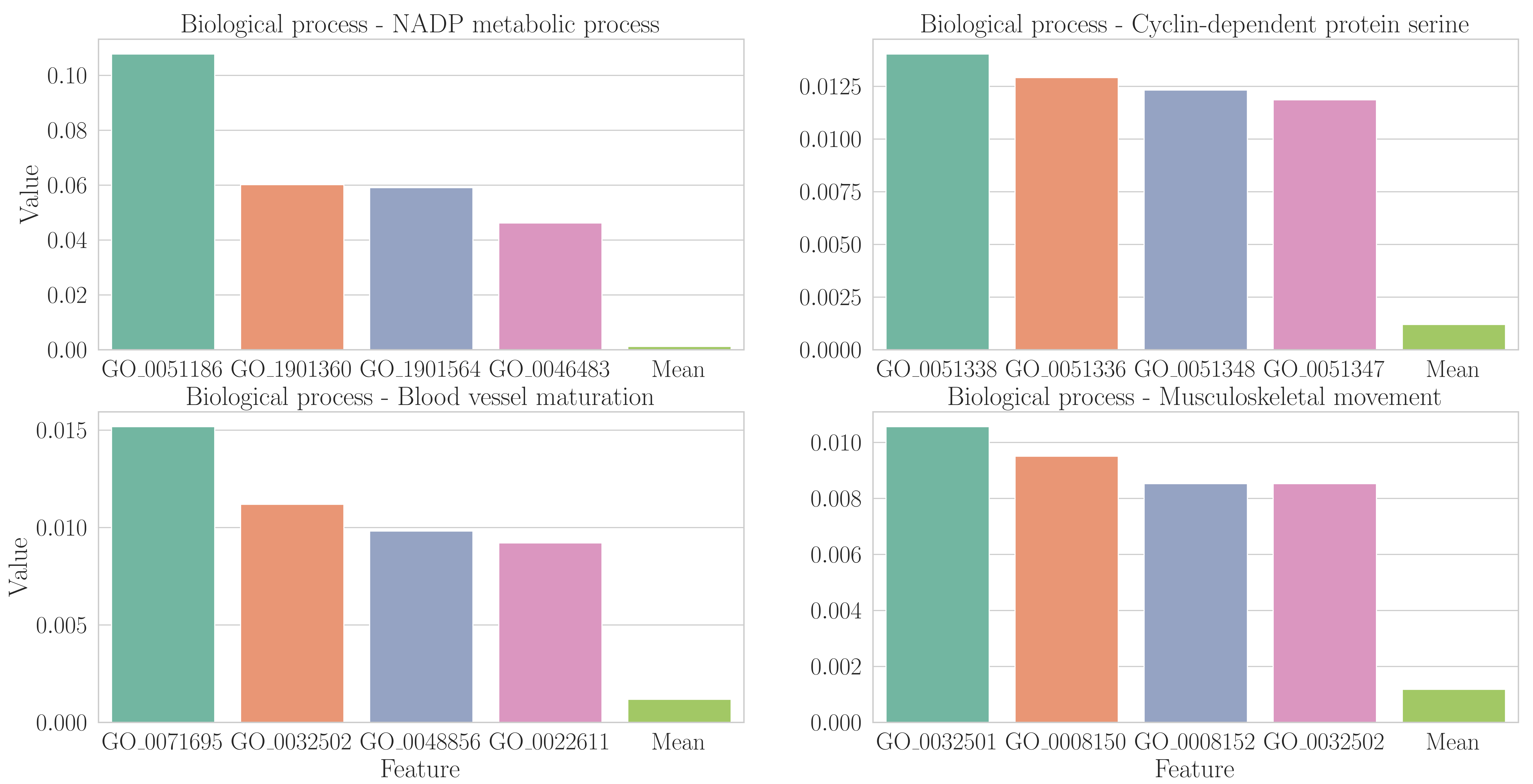}
  \caption{Examples of local explanations for four missing edges with the highest score. The sum of all feature values for an edge is equal to the edge's score.}
  \label{fig:exp-local}
\end{figure*}

The local interpretation of top four recommendations for the biological process node of the same year are shown in Figure~\ref{fig:exp-local}. For each of the four recommendations, we show four features that contribute the most, the amount they contribute, and the mean value of features in this vector. Overall, we can see that in our examples some features stand out, but even the top feature is in some cases only ten times higher than the mean value. The feature that stands out most in these examples is \emph{GO\textunderscore 0051186 (cofactor metabolic process)} between nodes that represent terms \emph{GO\textunderscore 0008150 (biological process)} and \emph{GO\textunderscore 0006739 (NADP metabolic process)}. This feature has value $0.11$ while the second highest \\\emph{GO\textunderscore 1901360 (organic cyclic compound metabolic process)} has $0.06$. 

\begin{figure*}[h!]
  \centering
  \includegraphics[width=\linewidth]{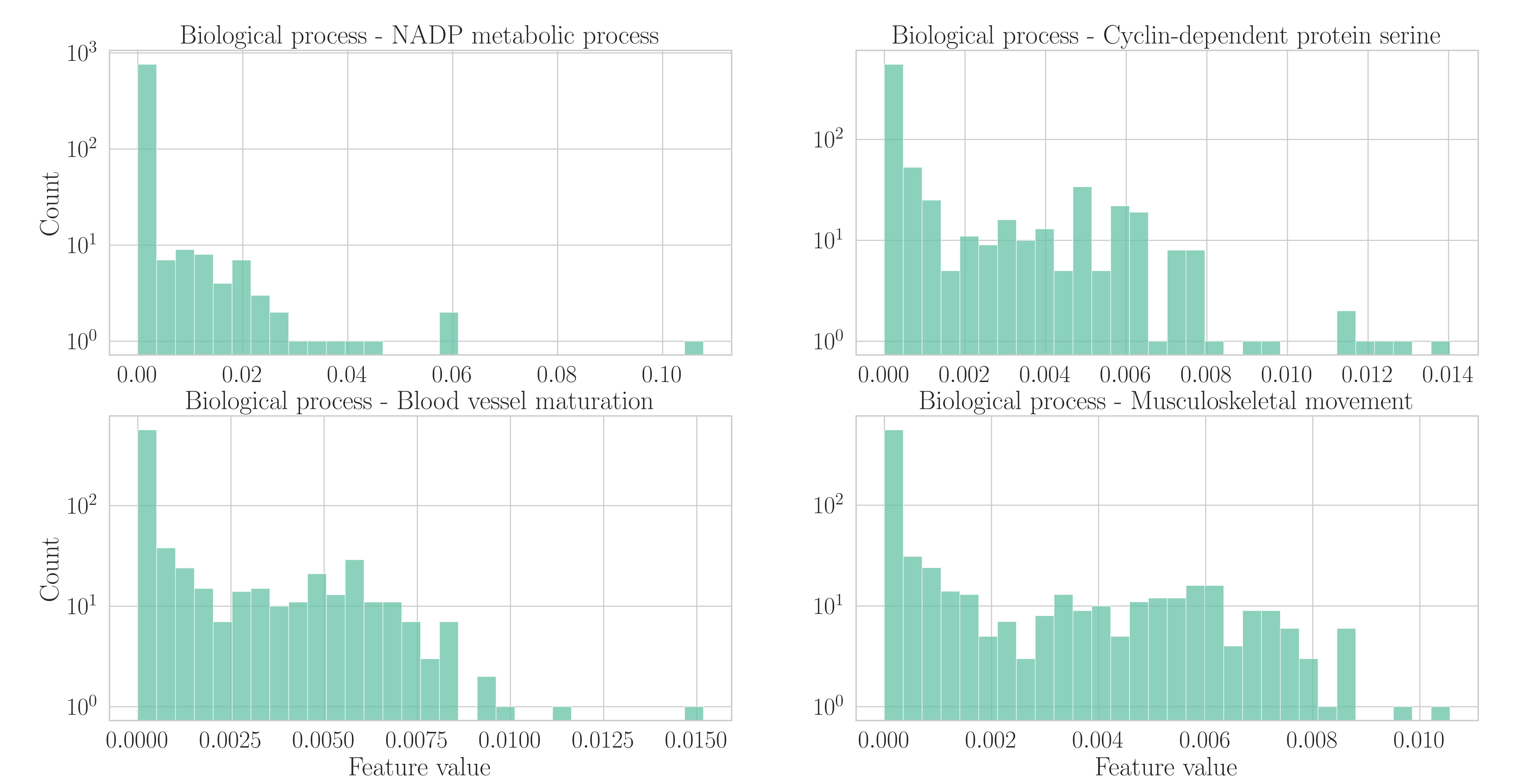}
  \caption{Distribution of feature values for local explanations. The x axis shows the value of the feature, while the y axis (logarithmic scale) represents the number of features with the value.}
  \label{fig:exp-dist}
\end{figure*}

Further, we can look at the distribution of feature values in the local interpretation and determine whether some of these values stand out. An example of this can be seen in figure~\ref{fig:exp-dist}, where distribution of feature values of top four recommendations are shown. We can see that most features of each recommendation are zero, while only a few really stand out.

\section{Discussion}
This section summarises our work and discusses the main advantages and disadvantages of the proposed approach for finding missing and redundant edges in ontologies.

The main goal of our approach is to generate reasonable predictions for missing and redundant links in ontologies solely based on their structure. This is done by scoring edges using link prediction algorithms. In Section~\ref{sec:results-link} we empirically show the results of link prediction on graphs we get by transforming ontologies. We see that link prediction works well on graphs with a high number of average edges per node but badly on graphs whose structure resembles a tree. This means that our methodology will be much more reliable on graphs with many edges and unreliable on ontologies whose structure resembles a tree.

In Section~\ref{sec:results-candidates} we tested our approach by using different versions of the same ontology and testing whether the created candidates occur in other versions. We do this by creating candidates for missing edges on the version $t$ of the ontology and checking their occurrence on version $t+1$. The results for the predicted missing edges in Table~\ref{tab:gene-missing} show that the candidates with the higher score have a better chance to occur in the following version, while the redundant edges occur with roughly the same frequency regardless of their score.

Since the results show that a significant percentage of candidates for new edges occurs in the following version of the ontology, this methodology could help annotate larger ontologies where connections can easily be missed. In practice, the project coordinator could set up a web application where the annotator would select nodes he/she is interested in and get the missing edges with the highest, and existing edges with the lowest score. To further increase the possibility of correct annotation, explanations for the relevant edges could also be shown. The main bottleneck for our approach is the calculation of the embedding, which can be done offline. This means that creating recommendations for annotators would be fast and would not need a lot of resources -- the developed approach could serve as an on-line ontology annotation assistant.  

There are a few disadvantages to the proposed methodology. The biggest is that the methodology works well on ontologies where each node has many connections but poorly on the ones whose structure resembles a tree. Such ontologies probably do not contain enough information to recommend new connections only based on their structure. A possible solution could be to include additional information such as the description of classes and recommend edges based on this semantic information.

Another disadvantage is that the approach needs quadratic space to store scores for each connection. This could prove problematic for large ontologies where such an approach is needed even more due to the number of possible connections that can easily be missed. In practice, this is not necessarily problematic since embeddings are usually small enough to fit inside the memory and can be used to calculate scores for only a subset of $k$ nodes. This lowers the space complexity to $|N|\cdot k$, which can easily fit inside the memory and gives the same results.  

We note that this work primarily focuses on outlining a scalable end-to-end methodology for graph-based link prediction on ontologies. As such, there exist possible improvements to many of the approaches adopted as part of the pipeline.

For example, we demonstrate how KG specific methods compare to
methods that operate on simple graphs.
However, there exist even more specialized methods, such as \linebreak ontology-specific embeddings~\cite{smaili2018onto2vec, smaili2018opa2vec, chen2020owl2vec, chen2018on2vec, kulmanov2019el}. These approaches are tailored to capture the higher expressivity that ontologies offer compared to knowledge graphs and usually utilize lexical information (meta-data) about nodes and relations, which we ignore. Leveraging this additional information would likely produce better results.

Other choices could also be made when it comes to ontology pre-processing and the ontology-to-graph conversion step. These include extending given ontologies with related ones, ontology pruning, entailment reasoning before conversion and the choice of ontology-to-graph conversion protocol itself.

\section{Conclusions}
In this work, graph-based machine learning approaches were used in the proposed methodology for finding missing and redundant edges in ontologies. We showed that this approach yields good results on larger ontologies when nodes have a high average degree.

The proposed approach could prove useful for annotators of large ontologies or domain experts (e.g., biologists) to find the connections that are the most likely to belong in the ontology. 

In further work, we plan to collaborate with domain experts to further analyze the performance of our methodology in a real-life setting.
We also intend to study different approaches for pre-processing ontologies and representing them as graphs, since this is one potential area where incorporating more of the available semantic information into the model could help improve results.
Finally, we wish to move our focus to ontology-specific embeddings and other methods that utilize meta-data about entities. We suspect that taking full advantage of these additional features can significantly improve the results and further explore the limits of the presented methodology.

\section*{Acknowledgments}

This work was supported by the European Union's Horizon 2020 research and innovation programme under grant agreement 863059 (FNS-Cloud, Food Nutrition Security). The work of the last author was supported by the Slovenian Research Agency (Young researcher grant).

We would like to thank Barbara Koro\v{s}i\v{c} Seljak, Tome Eftimov, and Gjorgjina Cenikj for providing valuable early feedback about the direction of our work.

\bibliographystyle{sm}
\bibliography{sm}

\end{document}

%% file: table1.tex
\begin{tabular}{>{\centering\arraybackslash}m{5.9cm}
>{\centering\arraybackslash}m{2cm}
>{\centering\arraybackslash}m{2cm}
>{\centering\arraybackslash}m{2cm}
>{\centering\arraybackslash}m{5.2cm}}

    \hline
    Article Title & Article type & Considered data & Method type & Description \\
    \hline

Onto2Vec: joint vector-based representation of biological entities and their ontology-based annotations~\cite{smaili2018onto2vec}
& method & Gene ontology  & syntactic ontology embedding    & Represents ontology axioms as sentences and trains a word2vec model to generate embeddings. Evaluated on a PPI task. \\

OPA2Vec: Combining formal and informal content of biomedical ontologies to improve similarity-based prediction~\cite{smaili2018opa2vec}
& method   & PhenomeNET and Gene ontology & syntactic ontology embedding  & Extends Onto2Vec by including informal information such as class descriptions in the axiom sentences.\\

On2Vec : Embedding-based Relation Prediction for Ontology Population~\cite{chen2018on2vec} 
& method & Yago, ConceptNet and DBPedia ontology    & graph-based ontology embedding    & Adapts translational KG embeddings for ontologies by accounting for hierarchical relations. Evaluated on a relation prediction task.   \\

OWL2Vec*: Embedding of \linebreak OWL Ontologies~\cite{chen2020owl2vec} 
& method  & HeLiS, FoodOn, and Gene ontology  & hybrid ontology embedding    & Combines concepts from OPA2Vec with biased random walks on the ontology graph.  \\

\hline

Predicting Gene-Disease Associations with Knowledge Graph Embeddings over Multiple Ontologies~\cite{nunes2021genedisease}
& application    & Several biomedical ontologies       & various KG and ontology embeddings    & Combines several biomedical ontologies along with annotations and applies several embeddings for the task of gene-disease prediction.   \\

Predicting Candidate Genes From Phenotypes, Functions, And Anatomical Site Of Expression~\cite{chen2020candidategenes} 
& application and a method & Several biomedical ontologies      & various ontology embeddings  & Combines data from several biomedical ontologies. Presents a novel domain-specific embedding model and evaluates it against existing ontology embeddings on a gene-disease prediction task. \\

Ontology-based prediction of \linebreak cancer driver genes~\cite{althubaiti2019cancergenes}
& application     & Several biomedical ontologies    & syntactic ontology embeddings    & Combines data from several biomedical ontologies, generates features with OPA2Vec and trains a model to predict cancer genes.                                                  \\

Gene function prediction based on Gene Ontology Hierarchy Preserving Hashing~\cite{zhao2018genefunc}
& application and a method & Gene ontology  & novel semantic similarity method   & Gene ontology terms are represented by a hierarchy-preserving hash function before computing semantic similarity for gene-function prediction.  \\

Protein–protein interaction inference based on semantic similarity of Gene Ontology terms~\cite{zhang2016ppi}
& application & Gene ontology    & various semantic similarity methods   & Integrates multiple semantic similarity measures to improve PPI prediction on the Gene ontology. \\

\hline

Semantic similarity and machine learning with ontologies~\cite{kulmanov2020ontosurvey}
& survey   & Gene ontology  & various SSM, simple graph embeddings, KG embeddings and ontology embeddings & Survey on ML with ontologies. Covers both traditional SSM and recent ontology embedding methods.

\end{tabular}%

%% file: table2.tex
\begin{tabular}{|c|c|c|c|}
    \hline
    Ontology    & |N|   & |E|   &  Components   \\ \hline
    Marine~\cite{tzitzik2013marineTLO}  &  108  &  156  &    2   \\
    Anatomy~\cite{Bard2012AEO} &  249  &  366  &    1   \\
    SCTO~\cite{El-Sappagh2018SCTO}&  321  &  370  &     1   \\
    Emotions~\cite{Hastings2011Emotions}  &  631  &  773  &   1   \\
    EHDAA2~\cite{Bard2012EHDAA2} & 2743 & 12894 &   15 \\
    Food~\cite{Dooley2018FoodOn}        & 28740 & 35897   &  107  \\
    LKN~\cite{Ramsak2018lkn}          & 20011 & 68503 &   2427 \\
    GO~\cite{Ashburner2000GO} & 44167 & 101504 &   1   \\ \hline
\end{tabular}

%% file: table3.tex
\begin{tabular}{lllllll}

    \hline
    Ontology & Classes & Individuals & Object properties & Subsumption axioms & Restrictions & Set axioms \\ \hline
    Marine & 104 & 3 & 92 & 105 & 0 & 0 \\
    Anatomy & 250 & 0 & 11 & 366 & 101 & 0 \\
    SCTO & 394 & 18 & 8 & 341 & 251 & 111 \\
    Emotions & 688 & 36 & 29 & 774 & 94 & 40 \\
    EHDAA2 & 2734 & 0 & 9 & 13366 & 10283 & 0 \\
    Food & 45942 & 381 & 68 & 39155 & 8860 & 2543 \\
    GO & 62201 & 0 & 9 & 90583 & 30704 & 23493 \\
    \bottomrule
\end{tabular}

%% file: table4.tex
\begin{tabular}{lllllllll}
\toprule
Dataset       &              Marine           &             Anatomy           &                SCTO           &  Emotions                     &          EHDAA2               &              FoodOn           &  LKN                            &                GO           \\
Baseline      &                               &                               &                               &                               &                               &                               &                                 &                               \\
\midrule                                                                                                                                                                                                                                                                            
Adamic        &   61.01 ($\pm$ 3.79)          &  51.05 ($\pm$ 0.90)           &  56.22 ($\pm$ 2.59)           &  50.47 ($\pm$ 0.68)           &  71.88 ($\pm$ 0.46)           &  50.91 ($\pm$ 0.06)           &  62.83 ($\pm$ 0.50)             &  65.39 ($\pm$ 0.13)           \\
Jaccard       &   60.72 ($\pm$ 3.41)          &  51.02 ($\pm$ 0.92)           &  56.11 ($\pm$ 2.47)           &  50.47 ($\pm$ 0.68)           &  62.30 ($\pm$ 0.81)           &  50.91 ($\pm$ 0.06)           &  62.77 ($\pm$ 0.50)             &  64.95 ($\pm$ 0.11)           \\
Prefferential &   70.76 ($\pm$ 4.47)          &  52.31 ($\pm$ 4.30)           &  55.70 ($\pm$ 2.99)           &  52.34 ($\pm$ 3.32)           &  83.38 ($\pm$ 0.35)           &  47.54 ($\pm$ 0.46)           &  88.75 ($\pm$ 0.34)             &  69.53 ($\pm$ 0.16)           \\
GAE           &   63.73 ($\pm$ 5.18)          &  57.44 ($\pm$ 4.67)           &  58.76 ($\pm$ 4.28)           &  58.59 ($\pm$ 4.49)           &  80.38 ($\pm$ 0.67)           &  53.02 ($\pm$ 0.43)           &  83.80 ($\pm$ 2.99)             &  68.16 ($\pm$ 0.40)           \\
GAT           &   45.35 ($\pm$ 2.33)          &  54.50 ($\pm$ 3.63)           &  50.51 ($\pm$ 3.43)           &  51.98 ($\pm$ 1.36)           &  72.40 ($\pm$ 0.72)           &  54.26 ($\pm$ 1.39)           &  63.51 ($\pm$ 8.88)             &  77.75 ($\pm$ 0.21)           \\
GCN           &   61.75 ($\pm$ 4.92)          &  59.03 ($\pm$ 5.27)           &  54.73 ($\pm$ 2.33)           &  56.51 ($\pm$ 5.46)           &  69.69 ($\pm$ 0.66)           &  57.09 ($\pm$ 0.57)           &  75.69 ($\pm$ 1.27)             &  75.66 ($\pm$ 0.88)           \\
GIN           &   59.81 ($\pm$ 6.16)          &  59.47 ($\pm$ 4.62)           &  54.90 ($\pm$ 3.03)           &  54.69 ($\pm$ 3.81)           &  70.64 ($\pm$ 0.87)           &  58.11 ($\pm$ 0.18)           &  73.23 ($\pm$ 0.57)             &  77.19 ($\pm$ 0.19)           \\
SNoRe         &   70.79 ($\pm$ 2.05)          &  57.59 ($\pm$ 2.45)           &  \bfseries 59.06 ($\pm$ 3.23) &  60.47 ($\pm$ 2.83)           &  69.06 ($\pm$ 0.90)           & \bfseries  64.82 ($\pm$ 0.16) &  86.91 ($\pm$ 0.29)             & \bfseries  79.82 ($\pm$ 0.19) \\
node2vec      &   71.01 ($\pm$ 5.07)          &  53.20 ($\pm$ 3.54)           &  52.25 ($\pm$ 2.10)           &  47.71 ($\pm$ 2.04)           &  74.51 ($\pm$ 0.83)           &  51.25 ($\pm$ 0.44)           &  86.47 ($\pm$ 0.36)             &  76.37 ($\pm$ 0.11)           \\
metapath2vec  &   \bfseries 76.09 ($\pm$ 3.55)&  57.36 ($\pm$ 5.61)           &  41.42 ($\pm$ 3.16)           &  49.20 ($\pm$ 4.92)           &  78.93 ($\pm$ 0.39)           &  57.68 ($\pm$ 0.53)           &  76.91 ($\pm$ 0.47)             &  53.76 ($\pm$ 0.22)           \\
TransE        &   74.82 ($\pm$ 6.68)          &  56.23 ($\pm$ 3.16)           &  55.99 ($\pm$ 2.34)           &  54.16 ($\pm$ 1.24)           &  \bfseries 84.63 ($\pm$ 0.23) &  64.56 ($\pm$ 1.41)           &  \bfseries 89.62 ($\pm$ 0.29)      &  75.20 ($\pm$ 0.54)           \\
RotatE        &   75.98 ($\pm$ 5.20)          &  50.36 ($\pm$ 4.47)           &  55.69 ($\pm$ 2.99)           &  49.74 ($\pm$ 2.53)           &  71.75 ($\pm$ 0.90)           &  47.82 ($\pm$ 0.19)           &  88.68 ($\pm$ 0.38)             &  77.61 ($\pm$ 0.25)           \\
Spectral      &   43.48 ($\pm$ 10.15)         &  \bfseries 59.62 ($\pm$ 5.26) &  55.84 ($\pm$ 1.49)           &  \bfseries 61.50 ($\pm$ 2.56) &  68.32 ($\pm$ 2.23)           &  49.27 ($\pm$ 3.57)           &  83.93 ($\pm$ 0.94)             &  61.07 ($\pm$ 1.92)           \\

\bottomrule
\end{tabular}

%% file: table5.tex
\begin{tabular}{lllllllll}
\toprule
Dataset       &              Marine           &             Anatomy           &                SCTO           &  Emotions                     &          EHDAA2               &              FoodOn           &  LKN                            &                GO           \\
Baseline      &                               &                               &                               &                               &                               &                               &                                 &                               \\
\midrule                                                                                                                                                                                                                                                                            
Adamic        &  60.96 ($\pm$ 3.72)           &  51.52 ($\pm$ 0.72)           &  56.27 ($\pm$ 3.20)           &  50.68 ($\pm$ 0.60)           &  75.73 ($\pm$ 0.51)           &  50.91 ($\pm$ 0.06)           &   62.88 ($\pm$ 0.50)            &  65.39 ($\pm$ 0.13)           \\
Jaccard       &  59.54 ($\pm$ 2.71)           &  51.02 ($\pm$ 0.86)           &  55.33 ($\pm$ 2.22)           &  50.46 ($\pm$ 0.61)           &  55.48 ($\pm$ 0.73)           &  50.91 ($\pm$ 0.06)           &   62.12 ($\pm$ 0.48)            &  64.95 ($\pm$ 0.11)           \\
Prefferential &  72.44 ($\pm$ 5.63)           &  57.09 ($\pm$ 2.97)           &  \bfseries 66.10 ($\pm$ 2.72) &  64.29 ($\pm$ 3.53)           &  \bfseries 86.89 ($\pm$ 0.53) &  47.54 ($\pm$ 0.46)           &   89.68 ($\pm$ 0.33)            &  69.53 ($\pm$ 0.16)           \\
GAE           &  70.74 ($\pm$ 4.92)           &  59.61 ($\pm$ 4.65)           &  66.06 ($\pm$ 4.22)           &  \bfseries 65.24 ($\pm$ 3.62) &  84.76 ($\pm$ 0.76)           &  53.02 ($\pm$ 0.43)           &   83.80 ($\pm$ 2.99)            &  68.16 ($\pm$ 0.40)           \\
GAT           &  45.75 ($\pm$ 1.69)           &  55.29 ($\pm$ 1.07)           &  50.35 ($\pm$ 3.36)           &  52.24 ($\pm$ 1.80)           &  70.96 ($\pm$ 0.47)           &  54.26 ($\pm$ 1.39)           &  66.73 ($\pm$ 11.95)            &  77.75 ($\pm$ 0.21)           \\
GCN           &  60.15 ($\pm$ 4.62)           &  58.83 ($\pm$ 3.70)           &  56.11 ($\pm$ 2.83)           &  58.47 ($\pm$ 3.75)           &  68.29 ($\pm$ 1.05)           &  57.09 ($\pm$ 0.57)           &   75.69 ($\pm$ 1.27)            &  75.66 ($\pm$ 0.88)           \\
GIN           &  63.01 ($\pm$ 3.65)           &  58.44 ($\pm$ 3.39)           &  55.38 ($\pm$ 3.09)           &  57.30 ($\pm$ 3.25)           &  69.52 ($\pm$ 1.22)           &  58.11 ($\pm$ 0.18)           &   73.23 ($\pm$ 0.57)            &  77.19 ($\pm$ 0.19)           \\
SNoRe         &  70.80 ($\pm$ 0.68)           &  58.28 ($\pm$ 1.31)           &  61.67 ($\pm$ 3.35)           &  61.06 ($\pm$ 3.07)           &  65.95 ($\pm$ 1.15)           & \bfseries  64.82 ($\pm$ 0.16) &   86.06 ($\pm$ 0.54)            & \bfseries  79.82 ($\pm$ 0.19) \\
node2vec      &  73.97 ($\pm$ 3.99)           &  57.79 ($\pm$ 3.27)           &  53.05 ($\pm$ 1.38)           &  50.31 ($\pm$ 2.49)           &  72.80 ($\pm$ 1.08)           &  51.25 ($\pm$ 0.44)           &   87.19 ($\pm$ 0.39)            &  76.37 ($\pm$ 0.11)           \\
metapath2vec  &  \bfseries 78.76 ($\pm$ 4.79) &  61.41 ($\pm$ 6.01)           &  46.57 ($\pm$ 2.42)           &  57.53 ($\pm$ 5.23)           &  79.47 ($\pm$ 0.73)           &  57.68 ($\pm$ 0.53)           &   73.24 ($\pm$ 0.49)            &  53.76 ($\pm$ 0.22)           \\
TransE        &  74.94 ($\pm$ 4.54)           &  58.89 ($\pm$ 3.79)           &  64.10 ($\pm$ 3.40)           &  60.66 ($\pm$ 3.23)           &  74.55 ($\pm$ 0.85)           &  64.56 ($\pm$ 1.41)           &   \bfseries 91.80 ($\pm$ 0.20)  &  75.20 ($\pm$ 0.54)           \\
RotatE        &  75.83 ($\pm$ 4.11)           &  \bfseries 62.13 ($\pm$ 3.38) &  63.27 ($\pm$ 3.59)           &  62.18 ($\pm$ 1.54)           &  85.37 ($\pm$ 0.42)           &  47.82 ($\pm$ 0.19)           &   91.19 ($\pm$ 0.26)            &  77.61 ($\pm$ 0.25)           \\
Spectral      &  55.70 ($\pm$ 9.71)           &  61.05 ($\pm$ 3.79)           &  58.01 ($\pm$ 2.87)           &  59.25 ($\pm$ 2.07)           &  64.87 ($\pm$ 2.49)           &  49.27 ($\pm$ 3.57)           &   81.28 ($\pm$ 2.39)            &  61.07 ($\pm$ 1.92)           \\

\bottomrule
\end{tabular}

%% file: table6.tex
\begin{tabular}{ccc}

    \hline
    Year & Added & Removed \\ \hline
    2016 & 7492 & 3302  \\
    2017 & 7497 & 2960  \\
    2018 & 15905 & 13704  \\
    2019 & 3781 & 3746  \\
    2020 & 2969 & 4381  \\
    2021 & 12653 & 17114  \\ \hline  
    
\end{tabular}

%% file: table7.tex
\begin{tabular}{lllllll}
\toprule
k\textbackslash \textrm{year (t)} & 2015 &  2016 & 2017 & 2018 & 2019 & 2020  \\
\midrule
10  &  0.200  &  0.200  &  0.000  &  0.100  &  0.300  &  0.000 \\
100  &  0.060  &  0.070  &  0.040  &  0.040  &  0.080  &  0.000 \\
500  &  0.022  &  0.022  &  0.016  &  0.044  &  0.052  &  0.006 \\
\bottomrule
\end{tabular}

%% file: table8.tex
\begin{tabular}{lllllll}
\toprule
k\textbackslash \textrm{year (t)} &  2015 &  2016 & 2017 & 2018 & 2019 & 2020  \\
\midrule
10  &  0.000  &  0.000  &  0.000  &  0.100  &  0.000  &  0.000 \\
100  &  0.040  &  0.020  &  0.020  &  0.070  &  0.130  &  0.010 \\
500  &  0.024  &  0.010  &  0.028  &  0.044  &  0.048  &  0.028 \\
\bottomrule
\end{tabular}